\documentclass[letterpaper]{article} 
\usepackage{aaai23}  
\usepackage{times}  
\usepackage{helvet}  
\usepackage{courier}  
\usepackage[hyphens]{url}  
\usepackage{graphicx} 
\usepackage{natbib}  
\usepackage{caption} 
\usepackage{algorithm}
\usepackage{algorithmic}
\usepackage{amsmath}

\usepackage{newfloat}
\usepackage{listings}
\DeclareCaptionStyle{ruled}{labelfont=normalfont,labelsep=colon,strut=off} 
\floatstyle{ruled}
\newfloat{listing}{tb}{lst}{}
\floatname{listing}{Listing}
\nocopyright

\title{End-to-end Manipulator Calligraphy Planning via Variational Imitation Learning}
\author {
    Fangping Xie,
    Pierre Le Meur,
    Charith Fernando
}
\affiliations {
    jeff@avatarin.com, pierre@avatarin.com, charith@avatarin.com
}

\begin{document}

\maketitle

\begin{abstract}
Planning from demonstrations has shown promising results
with the advances of deep neural networks.
One of the most popular real-world applications is
automated handwriting using a robotic manipulator.
Classically it is simplified as a two-dimension problem.
This representation is suitable for elementary drawings,
but it is not sufficient for Japanese calligraphy or complex work of art
where the orientation of a pen is part of the user's expression.
In this study, we focus on automated planning of Japanese calligraphy
using a three-dimension representation of the trajectory
as well as the rotation of the pen tip,
and propose a novel deep imitation learning neural network
that learns from expert demonstrations
through a combination of images and pose data.
The network consists of a combination of variational auto-encoder,
bi-directional LSTM, and Multi-Layer Perceptron (MLP).
Experiments are conducted in a progressive way, 
and results demonstrate that the proposed approach 
is successful in completion of tasks for real-world robots,
overcoming the distribution shift problem in imitation learning. 
The source code and dataset will be public.

\end{abstract}

\section{Introduction}
Learning from demonstrations or imitation learning is a research paradigm
\cite{argall2009survey} in machine learning applied to robotics planning.
It allows machines to learn a complicated human-like policy
that would otherwise be impossible to create programmatically for use
in a reinforcement learning algorithm. 
A popular application in this field is handwriting planning,
which uses complex bio-mechanisms to create trajectories. 
Many researchers achieved a certain degree of success
in reproducing the writing trajectories created by human demonstrations.
Previous methods for western calligraphy include
Gaussian mixture models \cite{khansari2011learning},
hidden Markov models \cite{williams2006extracting},
inverse optimal control \cite{yin2016synthesizing},
or dynamic movement primitive \cite{luo2020generalized}.
Other studies focusing on Chinese calligraphy made use of generative adversarial nets (GAN) \cite{chao2018generative, chao2020lstm} or
differential evolution \cite{gao2019data} to reproduce simple strokes.

In this study, we propose a policy for a real-world manipulator
which is capable of planning the task of Japanese calligraphy.
The policy imitates the style to write a relatively complicated character
using a set of demonstrations from an expert.
Previous studies mainly rely on the simplified two-dimension robot systems
or devices such as tablets.
For Japanese calligraphy, it is straightforward
to consider the pen tip in three dimensions, including orientation.
A robotic system using a manipulator with six degrees of freedom for the pen tip
is implemented in this work.
Along with the trajectory input, a third-person view camera is
introduced to capture the vision states.
Inspired by the successful imitation learning methods published in \cite{suzuki2021air}
for the application of bi-manual manipulation,
we propose a planner as a variational imitation learning network, which
yields significant performance experimentally.

\section{Preliminaries}
Handwriting planning can be described 
as a standard Markov decision process (MDP) model 
\begin{math}
\left \langle \textbf{\textit{S}}, \textbf{\textit{A}}, r, T \right \rangle
\end{math}, where \textbf{\textit{S}} and \textbf{\textit{A}} are the sets of feasible state and action, respectively. 
\begin{math} r\left(s, a\right) : \textbf{\textit{S}} \times \textbf{\textit{A}} \to \textbf{R} \end{math}
is the reward function for taking an action \textit{a} at state \textit{s},
while \begin{math} T \left( s' |s,a \right): \textbf{\textit{S}} \times \textbf{\textit{A}} \times \textbf{\textit{S}} \to \left[ 0, 1 \right] \end{math}
donates the dynamics of the environment as transition probability distribution from state \textit{s} to the next state \textit{s'} when action \textit{a} is taken. 
In imitation learning, the objective is to learn a policy \begin{math} \pi(a|s) : \textbf{\textit{S}} \times \textbf{\textit{A}} \to [0, 1] \end{math} defined as the probability of choosing action \textit{a} at state \textit{s} from demonstrations
\begin{math} \left\{ \tau(i) = (s_0 , a_0 , s_1 , a_1 , \cdots ) \right\}^{n}_{i=0} \end{math}
of an expert policy \begin{math} \pi_e \end{math}, where \textit{n} is the total number of states for the task.
We parameterize \begin{math} \pi \end{math} using a neural network with a set of parameters 
\begin{math} \theta \end{math}, noted as \begin{math} \pi_\theta \end{math}.

From demonstrations, we can expect the planning decision made by an expert on an MDP
\begin{math}
\left \langle \textbf{\textit{S}}_e, \textbf{\textit{A}}_e, r, T \right \rangle
\end{math} to be optimal for completing the task, e.g. Japanese calligraphy. 
This expert policy \begin{math} \pi_e \end{math}, based on the state representation can be parameterized by a learner policy \begin{math} \pi_\theta \end{math} through training a neural network with a specifically defined objective function, which is analog to the reward in the MDP. 

In general, for handwriting robotic manipulation,
a primitive planner corresponding to a parameterized policy that
maps states to actions \begin{math} \pi(a|s) \end{math} is integrated as a low-level controller.
In another word, given a primitive planner and the initial state
\begin{math} {s_0} \end{math},
The presented framework is expected to learn a high-level planning policy
\begin{math} \pi(s_{t+1} | s_t) \end{math} to produce a sequence of states
that will be further used by an inverse kinematics (IK) solver to generate action sequences
\begin{math} (a_1 , a_2 , \cdots ) \end{math} for robot movement control, as shown in Figure \ref{figwritemodel}.

\section{Proposed Approach}

\begin{figure*}[t]
\centering
\includegraphics[width=0.9\textwidth]{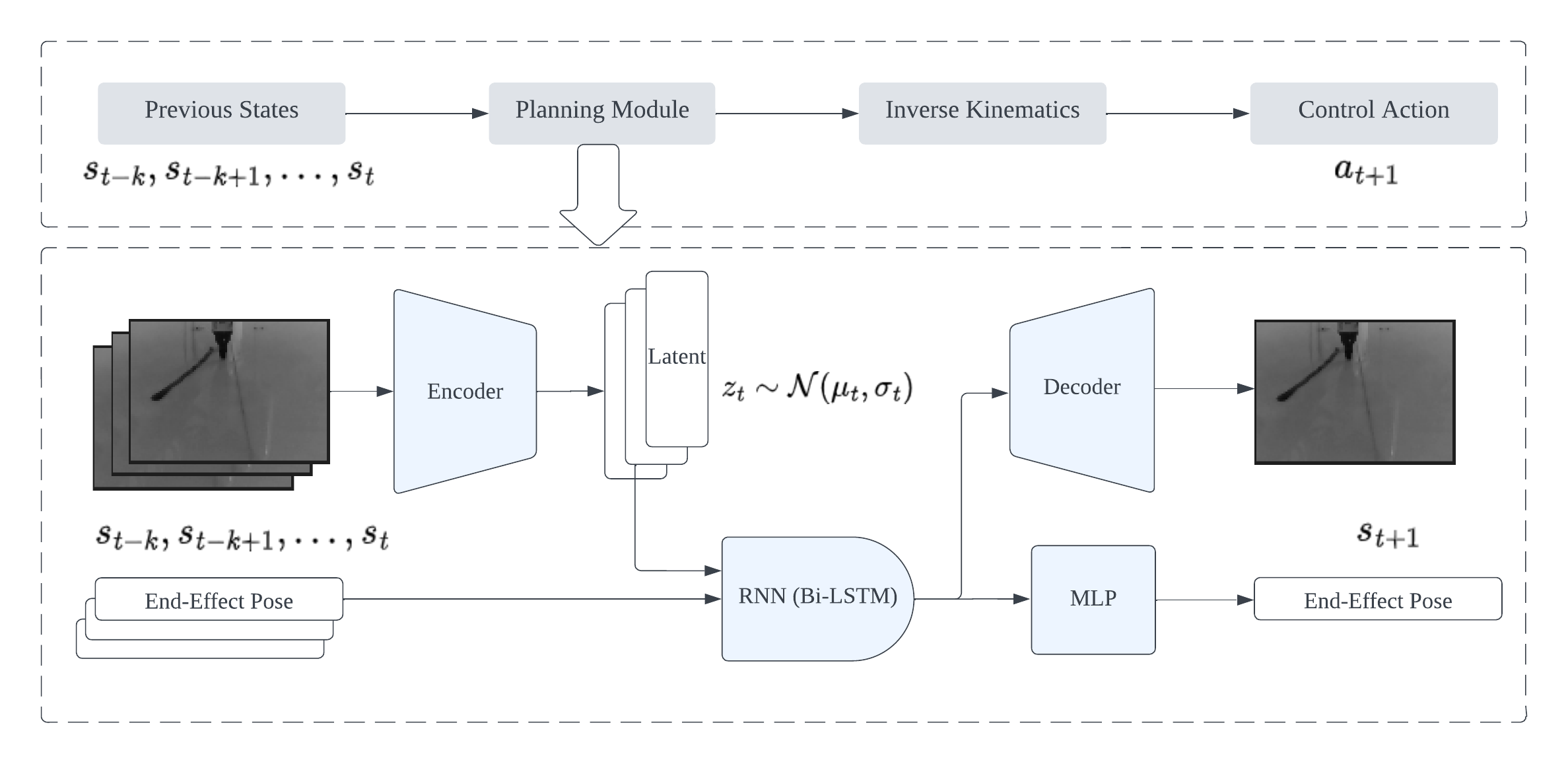} 
\caption{Planning pipeline. The top chart shows the procedures of the manipulator system, while the bottom one describes the planning module as a variational imitation learning model.}
\label{figwritemodel}
\end{figure*}

Provided with expert demonstrations \begin{math} \tau_{i=1}^{n} \end{math},
the target of imitation learning is to learn a policy \begin{math} \pi(a|s) \end{math}
that replicates the planning behavior of an expert. 
The proposed flowchart is described in Figure \ref{figwritemodel}.
The policy is decomposed into two modules, a probabilistic encoder
\begin{math} p_{\theta}(z|s) \end{math},
and a predictive decoder \begin{math} p_{\theta}(s'|z) \end{math}. We present the variational approximate posterior
\begin{math} q_{\phi}(z|s) \end{math}
as multivariate Gaussian with a diagonal covariance structure
as introduced in \cite{kingma2013auto}.

\begin{equation}\label{equalq}
\mathrm{log}\ q_\phi(z|s^{i}) = \mathrm{log}\ \mathcal{N}\left(z; \mu^{i}, \sigma^{2(i)}\mathrm{I}  \right)
\end{equation}
where the mean \begin{math} \mu^{i} \end{math}
and standard deviation \begin{math} \sigma^{i} \end{math} of the approximate posterior
are the outputs of the encoding network with variational parameters.

The second part of the policy decoder consists of two different modules, reconstruction of the predicted image and the pose states,
the latter serves as the action in the definition. 
Policy \begin{math} \pi_\theta \end{math} is learnt
by training an end-to-end neural network using Algorithm \ref{alg:algorithm}.

\subsection{Objective}
Following the variational model definition in \cite{kingma2013auto},
we define the training objective as
\begin{equation}\label{lossVIM}
\begin{split}
\mathcal{L}(\theta; s^{i})\simeq \frac{1}{2}\sum_{j=1}^{J}(1+\mathrm{log}((\sigma_j^i)^2)-(\mu_j^i)^2 - (\sigma_j^i)^2) \\
+\frac{1}{L}\sum_{l=1}^{L}\mathrm{log}\ p_\theta(s^i|z^{(i,l)})
\end{split}
\end{equation}
where \begin{math} z^{(i,l)}=\mu^i + \sigma^i \ast \epsilon^l, \mathrm{and}\ \epsilon^l\sim \mathcal{N}(0,\mathrm{I})\end{math}.
The first term is the estimated KL divergence of the approximate posterior from the prior, which acts as a regularizer, while the second term is the reconstruction error of predicted states from the decoder module, noted as \begin{math}
\mathcal{L}_2(\theta, s^i)
\end{math}.

Inspired by \cite{zhang2018deep}, for the second term in Equation \ref{lossVIM}, instead of a negative log-likelihood loss, a combination of mean absolute loss and mean square loss is applied as follows.
\begin{equation}\label{lossBC}
\begin{split}
\mathcal{L}_2(\theta, s^i) = \lambda_1\mathcal{L}_{MAE}(t) + \lambda_2\mathcal{L}_{MSE}(t)\\
\lambda_3\mathcal{L}_{MAE}(R) + \lambda_4\mathcal{L}_{MSE}(R)\\
+ \lambda_5\mathcal{L}_{MSE}(I)
\end{split}
\end{equation}
where \begin{math} t \end{math} stands for 3-state translation,
\begin{math} R \end{math} is the rotation which is represented by a 4-state quaternion, and \begin{math} I \end{math} stands for the image. Hyper-parameters \begin{math} \lambda_1, \lambda_2, \cdots, \lambda_5 \end{math}
leverages the prediction of the expert
actions and conciseness about the visual observations.
An extra weight \begin{math} \lambda_6 \end{math} is given for the KL divergence term in Equation \ref{lossVIM}.

\subsection{Residual Connection Feature Pyramid Network}

\begin{figure}[t]
\centering
\includegraphics[width=1.0\columnwidth]{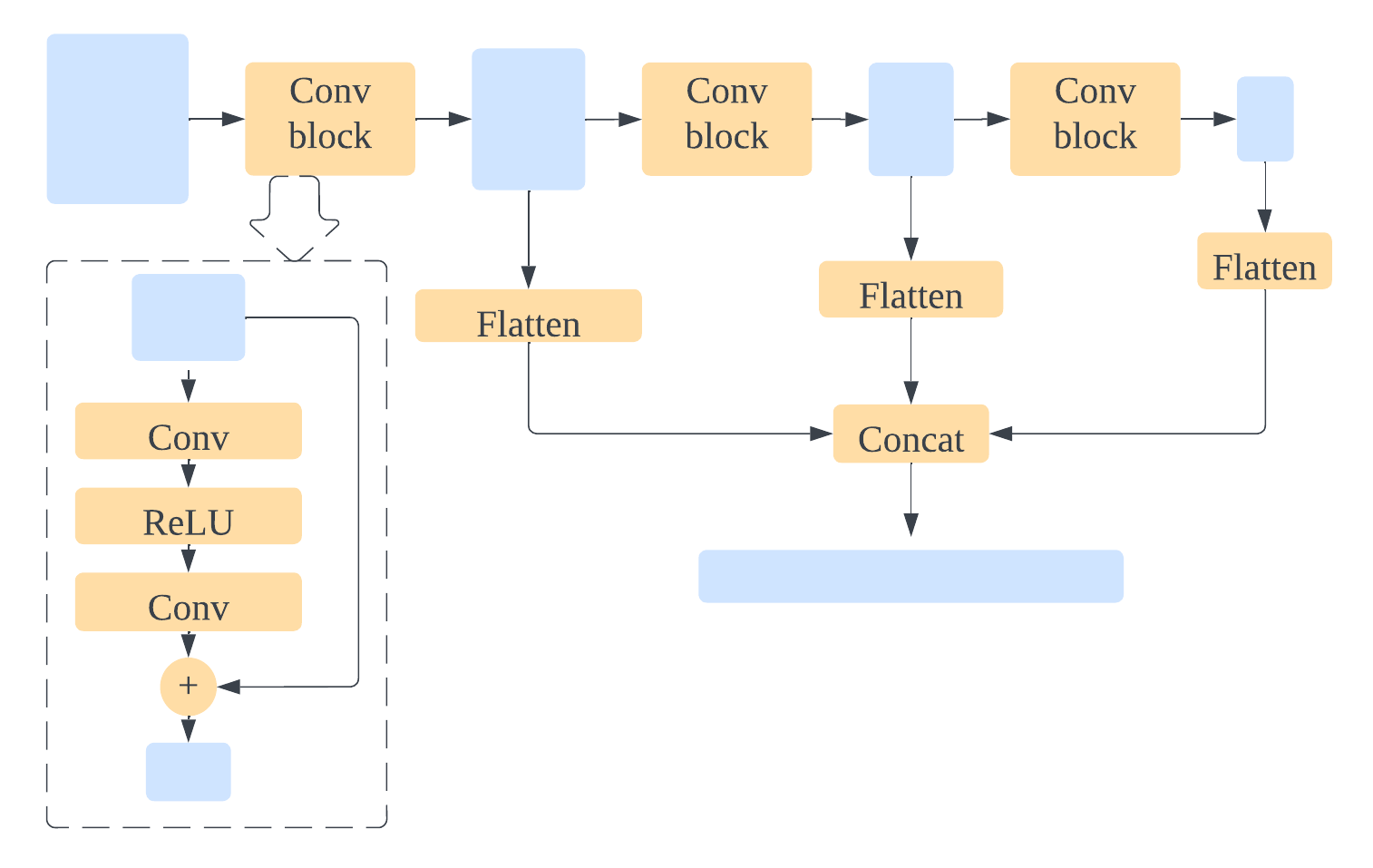} 
\caption{Residual connection feature pyramid network.}
\label{figFPN}
\end{figure}

We propose a residual connection feature pyramid network for the encoder module as described in Figure \ref{figFPN}. Resnet \cite{he2016deep} stands for the residual network. It is popular due to its success in the ImageNet image classification challenge, as well as its portability to other computer vision tasks such as object detection, segmentation, etc. Resnet architecture is based on residual blocks. A residual block uses a series of convolution layers along with a skip connection creating a direct link between the input and output of the block. The skip connection helps to reduce the problem of vanishing gradients in deep neural networks, as well as reducing the impact of less useful convolutional blocks.

Despite the advantages of Resnet, it faces the drawback of losing global image features after each convolutional block. The architecture of Feature Pyramidal Network (FPN) \cite{lin2017feature} with lateral connections is proposed to extract high-level semantic feature maps at all scales under marginal extra computational cost. Experimental results in various computer vision applications demonstrate the success of this architecture \cite{zhao2019object}. To the best of our knowledge, this is the first time applying FPN 
in the imitation learning from demonstrations.

\subsection{Latent Space Prediction}

For sequential data, the Long Short Term Memory (LSTM) architecture,
one type of Recurrent Neural Network (RNN),
achieves big success in tasks like handwriting recognition \cite{agmls2009novel} or speech recognition \cite{graves2013speech}
by resolving the issue related to gradient propagation through time, i.e. backpropagated error either explosion or decays exponentially.

Bi-directional LSTM (Bi-LSTM) was proposed by \cite{cornegruta2016modelling} and
it basically consists of two LSTMs: one taking the input in a forward direction,
and the other in a backward direction.
In another word, the output will contain information on both the backward and forward dynamics of the motion.

Planning in latent space \cite{NEURIPS2018_2de5d166, hafner2019dream}
attracts wide attention in the controlling and skills-learning field.
Inspired by the idea of latent imagination from images,
we use the decoder as a reconstruction of predicted images from the output of Bi-LSTM.

\subsection{Data Augmentation}
The planning model targets learning from expert demonstrations. In general, it is time-consuming and expensive to collect data from an expert.
Dataset is expanded with data augmentation by adding Gaussian noises to the pose data. Data augmentation also acts as a regularizer and reduces overfitting during model training.

Image augmentation has been proven in improving the performance
of methods learning from pixels \cite{kostrikov2020image}.
In particular, shift transformation is highly recommended for deep reinforcement learning. Besides, random noises in brightness, hue, and saturation are introduced to improve the system performance, especially in eliminating the impact of environment changes between the recording and inference times.



\begin{algorithm}[tb]
\caption{Variational imitation learning training}
\label{alg:algorithm}
\textbf{Input}: Expert trajectories \begin{math} \tau_i \sim \pi_e \end{math}, initial policy \begin{math} \pi \end{math} with parameters \begin{math} \theta \end{math}
\begin{algorithmic}[1] 
\WHILE{parameters \begin{math} \theta \end{math} not converged}
\STATE Sample batch \begin{math} S_{i,i+1} \end{math} from trajectory \begin{math} \tau_i \end{math}
\STATE Encode input \begin{math} S_i \end{math} to stochastic latent space
\begin{math} z_i \sim q_\theta(z_i|s_i) = \mathcal{N}(\mu_i, \sigma_i) \end{math}
\STATE Random sample \begin{math} \epsilon \end{math} from normal distribution
\begin{math} \mathcal{N}(\textbf{0, I})) \end{math}
\STATE Reparameterize \begin{math} z_i = \mu_i + \sigma_i * \theta \end{math}
\STATE Predict the next latent space \begin{math} z_{i+1} \end{math}
\STATE Decode latent space \begin{math} z_{i+1} \end{math}
to a probabilistic distribution \begin{math} p_\theta(s_{i+1} | z_{i+1}) \end{math}
\STATE Gradients \begin{math} g \gets \nabla \mathcal{L}\left(\theta, {S_{i,i+1}}\right) \end{math} (Eq. 1)
\STATE Update parameters \begin{math} \theta \end{math} of policy \begin{math} \pi \end{math} with gradients \begin{math} g \end{math}
\ENDWHILE
\STATE \textbf{return} policy \begin{math} \pi_\theta \end{math}
\end{algorithmic}
\end{algorithm}

\section{Experiments and Results}

Calligraphy learning starts from simple strokes to complicated characters consisting of dozens of strokes.
Furthermore, the spirit of calligraphy lies in the specified way of initializing and finishing each stroke and the movement speed in between.
These key factors determine the styling of calligraphy.
This work focuses on solving the shape reproduction of Japanese characters
from an expert which is the first step of automated planning of calligraphy.


\subsection{Inference Pipeline}

It is widely acknowledged \cite{ross2011reduction} that the evaluation of
imitation learning policies on pre-recorded datasets is biased
since the real-world environment will have changing parameters such as light intensity,
light orientation, and so on.
On top of environment changes, output from trained model will diverge from the expert path,
which is also ignored for evaluation on pre-recorded dataset.
Those changes between real-time data and pre-recorded data are called distribution shift.
From experimental results, the inference on the pre-recorded data
shows good performance, as illustrated in row A of Figure \ref{figMainRes}.
Our inference on a real-world robot iteratively captures the real-time
pose data and images as the input.

\subsection{Real World Robot Application}
Evaluation of learned policies
focuses on the experimental results of completion \cite{sun2014robot}
due to the lack of benchmark and widely accepted metrics \cite{teo2002robotic}.

We carry out experiments in a progressive way.
For handwriting tasks, it is straightforward that
more strokes increase the difficulties to complete the task.
Roughly, strokes in Japanese characters consist of
simple straight lines, curve-style strokes, and a combination of
these two basic types.

\begin{figure}[t]
\centering
\includegraphics[width=1.0\columnwidth]{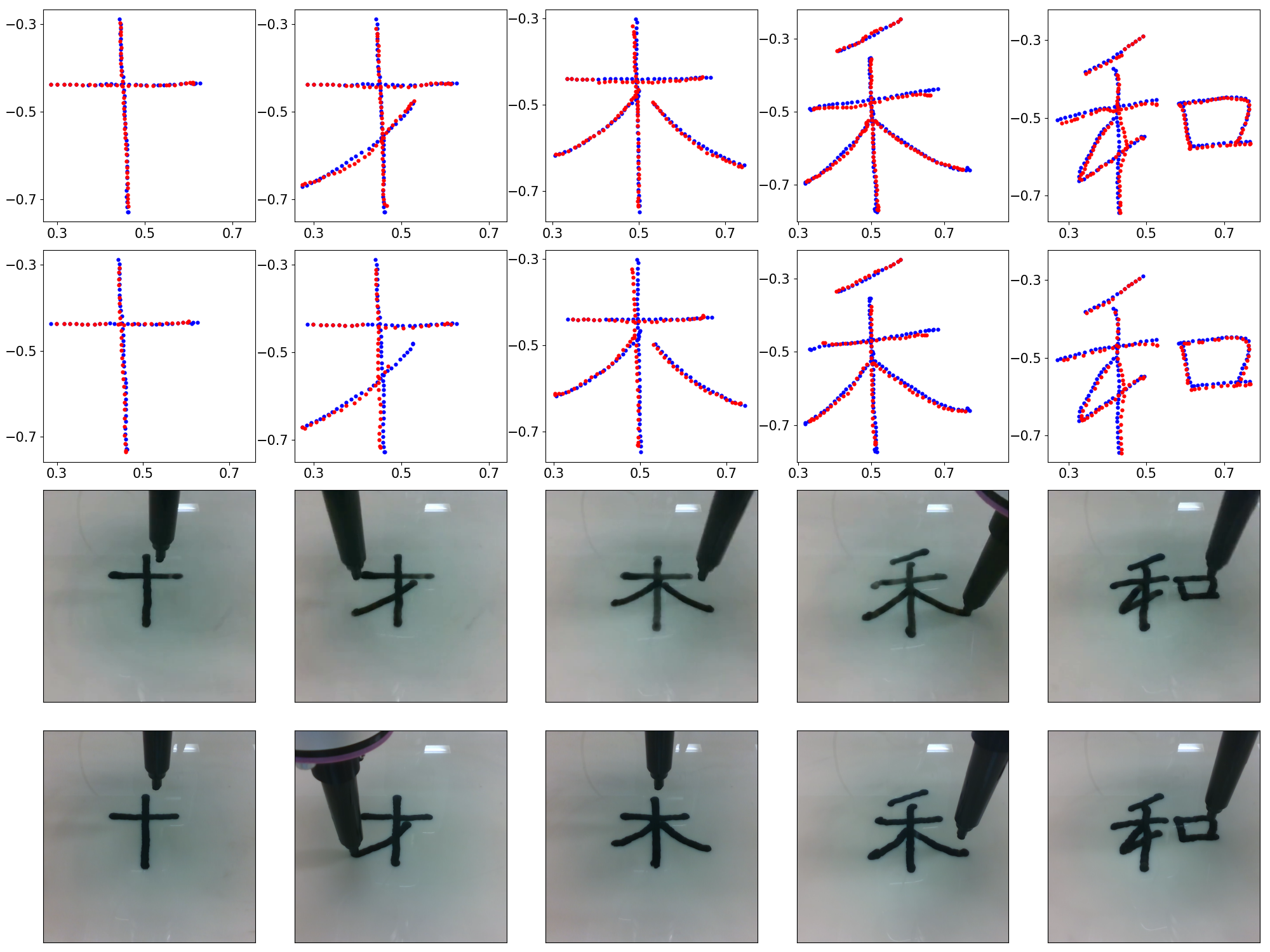} 
\caption{Inference results. First row shows the x-y plane plot of inference (red) on pre-recorded data compared with ground truth (blue), while second row is the inference (red) on the real-world robot xArm-7. Third and fourth rows are ground truth and real-world inference results, respectively.}
\label{figMainRes}
\end{figure}

Experiments are conducted on five different characters.
The proposed policy model learned from expert demonstrations
is capable to complete all these tasks.
In Figure \ref{figMainRes}, increasing number of strokes
is drawn in columns from left to right.
From the plots in second row, we show the correctness of the learned policy
on the shape reproduction of the real-world robot.
Comparing third and forth rows of the recorded images,
the shape during inference time remains the same writing style,
except for that some starting and ending parts of the strokes
have slight differences.


\subsection{Ablation Study}

Extensive ablation studies are performed to validate the
correctness and improvement of our approach.
All results shown in Figure \ref{figablation} are obtained through the same robot system and configuration on the most complicated Japanese character demonstrated in this work.
The challenge in behavior cloning is that
the method can complete the task on pre-recorded data,
but fails in real-time processing,
comparing top and bottom rows of Figure \ref{figablation}.
For expanded description of each result, please see
supplemental video.

\subsubsection{W/O Bidirectional LSTM}
Bi-LSTM is replaced with a standard LSTM layer in this experiment.
Comparing the result shown in first column of Figure
\ref{figablation}, the model is capable to complete the task with
slight loss of shape maintenance, e.g. the right part of the character.

\subsubsection{W/O Variational Encoder}
\begin{figure}[t]
\centering
\includegraphics[width=1.0\columnwidth]{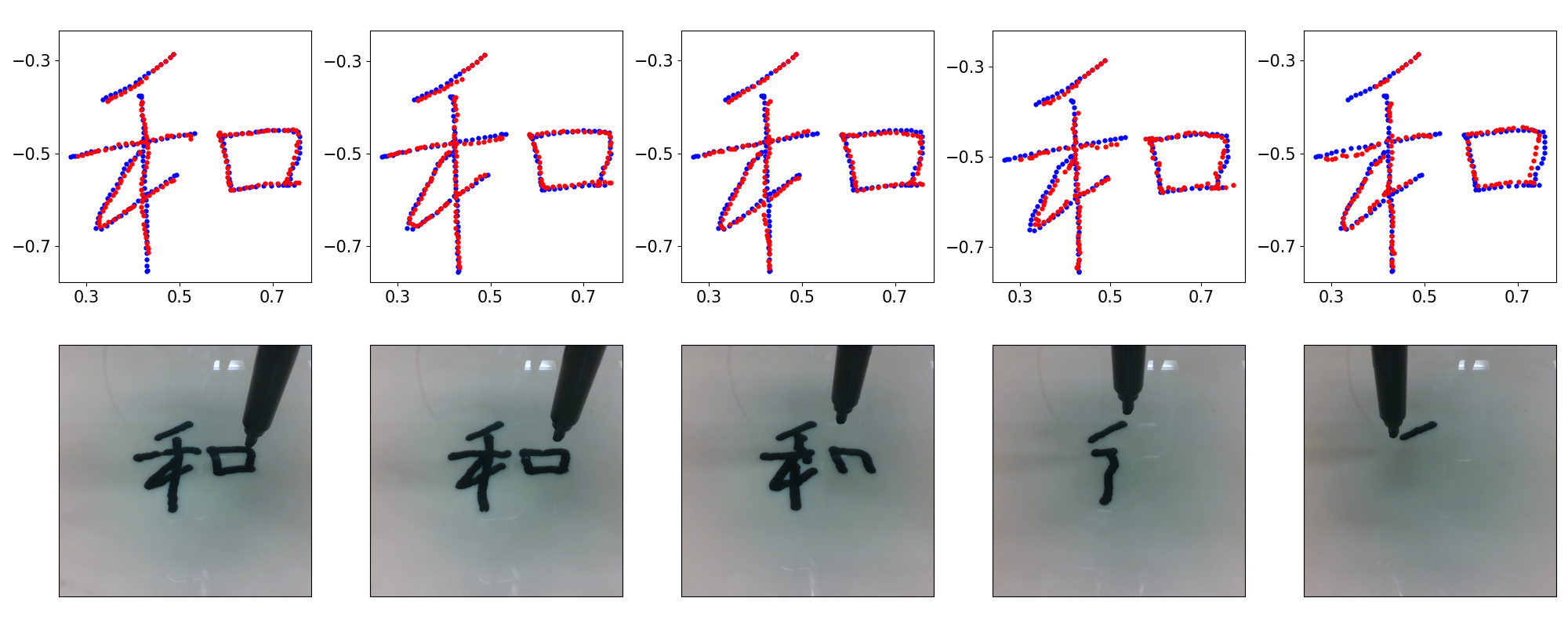} 
\caption{Ablation study.
Top row is the x-y plane plot of inference (red) on the recorded data compared with ground truth (blue)
on five experiments,
while bottom row is the corresponding real-world inference results.
}

\label{figablation}
\end{figure}

Similar to the previous section, the result in second column of Figure
\ref{figablation} without using
a variational encoder can also complete the task.
However, we can visualize the loss of straightness in some strokes.

\subsubsection{Resnet Encoder}
Figure \ref{figablation} third row shows the result of removing
the feature pyramid in the image encoding part, i.e. using Resnet encoder.
It is obvious that the learned policy has difficulties
to write the Japanese character properly, 
especially in the aspect of continuousness for some strokes.

\subsubsection{W/O Image Augmentation}

From forth column in Figure \ref{figablation}, 
the trained policy without using image augmentation
performs well on pre-recorded data (top),
but fails at the second stroke during real-time inference. 
The essential role of image augmentation
methods described in the approach section is verified in this experiment.

\subsubsection{W/O Pose Augmentation}
The result by training the policy without pose data augmentation, last column of Figure \ref{figablation}, shows that
the learned policy fails the task before the second stroke.
Pose data augmentation plays
a major role in our proposed method which absorbs the noise and
distribution shift of the system during the demonstration.

\section{Discussion}
We propose a variational imitation learning neural network as a policy for Japanese calligraphy planning
through learning from an expert.
The policy is decomposed into an encoder and decoder
to make use of the prediction of latent space.
This work targets solving the common distribution shift problem in imitation learning.
Experimental results on real-world robots demonstrate the
correctness of the proposed method in a progressive way.
Further ablation studies show the necessities and advantages of
each part in the completion of Japanese calligraphy tasks.

In this work, we focus on the 
three-dimension automated planning policy
for shape reproduction from an expert.
Further directions include training a policy that
is capable to perform multiple characters with external input,
e.g. speech instruction or robot and agent policies fusion.
Naturally, handwriting with a brush is a more complicated task,
which requires taking more states into consideration,
including but not limited to the velocity of pen tip movement,
the thickness of each stroke and force control.

\section{Acknowledgments}
This work was supported by grant JPMJMS2013,
Japan Science and Technology Agency Moonshot R\&D Program
as part of the project "Cybernetic being".

\appendix
\bibliography{aaai23}

\begin{thebibliography}{23}
\providecommand{\natexlab}[1]{#1}

\bibitem[{AGMLS, Bunke, and Schmiduber(2009)}]{agmls2009novel}
AGMLS, F.; Bunke, R.; and Schmiduber, J. 2009.
\newblock A novel connectionist system for improved unconstrained handwriting
  recognition.
\newblock \emph{IEEE Transactions on Pattern Analysis Machine Intelligence},
  31(5).

\bibitem[{Argall et~al.(2009)Argall, Chernova, Veloso, and
  Browning}]{argall2009survey}
Argall, B.~D.; Chernova, S.; Veloso, M.; and Browning, B. 2009.
\newblock A survey of robot learning from demonstration.
\newblock \emph{Robotics and autonomous systems}, 57(5): 469--483.

\bibitem[{Chao et~al.(2020)Chao, Lin, Zheng, Chang, Lin, Yang, and
  Shang}]{chao2020lstm}
Chao, F.; Lin, G.; Zheng, L.; Chang, X.; Lin, C.-M.; Yang, L.; and Shang, C.
  2020.
\newblock An LSTM Based Generative Adversarial Architecture for Robotic
  Calligraphy Learning System.
\newblock \emph{Sustainability}, 12(21): 9092.

\bibitem[{Chao et~al.(2018)Chao, Lv, Zhou, Yang, Lin, Shang, and
  Zhou}]{chao2018generative}
Chao, F.; Lv, J.; Zhou, D.; Yang, L.; Lin, C.-M.; Shang, C.; and Zhou, C. 2018.
\newblock Generative adversarial nets in robotic Chinese calligraphy.
\newblock In \emph{2018 IEEE International Conference on Robotics and
  Automation (ICRA)}, 1104--1110. IEEE.

\bibitem[{Cornegruta et~al.(2016)Cornegruta, Bakewell, Withey, and
  Montana}]{cornegruta2016modelling}
Cornegruta, S.; Bakewell, R.; Withey, S.; and Montana, G. 2016.
\newblock Modelling radiological language with bidirectional long short-term
  memory networks.
\newblock \emph{arXiv preprint arXiv:1609.08409}.

\bibitem[{Gao et~al.(2019)Gao, Zhou, Chao, Yang, Lin, Xu, Shang, and
  Shen}]{gao2019data}
Gao, X.; Zhou, C.; Chao, F.; Yang, L.; Lin, C.-M.; Xu, T.; Shang, C.; and Shen,
  Q. 2019.
\newblock A data-driven robotic Chinese calligraphy system using convolutional
  auto-encoder and differential evolution.
\newblock \emph{Knowledge-Based Systems}, 182: 104802.

\bibitem[{Graves, Mohamed, and Hinton(2013)}]{graves2013speech}
Graves, A.; Mohamed, A.-r.; and Hinton, G. 2013.
\newblock Speech recognition with deep recurrent neural networks.
\newblock In \emph{2013 IEEE international conference on acoustics, speech and
  signal processing}, 6645--6649. Ieee.

\bibitem[{Ha and Schmidhuber(2018)}]{NEURIPS2018_2de5d166}
Ha, D.; and Schmidhuber, J. 2018.
\newblock Recurrent World Models Facilitate Policy Evolution.
\newblock In Bengio, S.; Wallach, H.; Larochelle, H.; Grauman, K.;
  Cesa-Bianchi, N.; and Garnett, R., eds., \emph{Advances in Neural Information
  Processing Systems}, volume~31. Curran Associates, Inc.

\bibitem[{Hafner et~al.(2019)Hafner, Lillicrap, Ba, and
  Norouzi}]{hafner2019dream}
Hafner, D.; Lillicrap, T.; Ba, J.; and Norouzi, M. 2019.
\newblock Dream to control: Learning behaviors by latent imagination.
\newblock \emph{arXiv preprint arXiv:1912.01603}.

\bibitem[{He et~al.(2016)He, Zhang, Ren, and Sun}]{he2016deep}
He, K.; Zhang, X.; Ren, S.; and Sun, J. 2016.
\newblock Deep residual learning for image recognition.
\newblock In \emph{Proceedings of the IEEE conference on computer vision and
  pattern recognition}, 770--778.

\bibitem[{Khansari-Zadeh and Billard(2011)}]{khansari2011learning}
Khansari-Zadeh, S.~M.; and Billard, A. 2011.
\newblock Learning stable nonlinear dynamical systems with gaussian mixture
  models.
\newblock \emph{IEEE Transactions on Robotics}, 27(5): 943--957.

\bibitem[{Kingma and Welling(2013)}]{kingma2013auto}
Kingma, D.~P.; and Welling, M. 2013.
\newblock Auto-encoding variational bayes.
\newblock \emph{arXiv preprint arXiv:1312.6114}.

\bibitem[{Kostrikov, Yarats, and Fergus(2020)}]{kostrikov2020image}
Kostrikov, I.; Yarats, D.; and Fergus, R. 2020.
\newblock Image augmentation is all you need: Regularizing deep reinforcement
  learning from pixels.
\newblock \emph{arXiv preprint arXiv:2004.13649}.

\bibitem[{Lin et~al.(2017)Lin, Doll{\'a}r, Girshick, He, Hariharan, and
  Belongie}]{lin2017feature}
Lin, T.-Y.; Doll{\'a}r, P.; Girshick, R.; He, K.; Hariharan, B.; and Belongie,
  S. 2017.
\newblock Feature pyramid networks for object detection.
\newblock In \emph{Proceedings of the IEEE conference on computer vision and
  pattern recognition}, 2117--2125.

\bibitem[{Luo, Wu, and Gombolay(2020)}]{luo2020generalized}
Luo, Q.; Wu, J.; and Gombolay, M. 2020.
\newblock A Generalized Robotic Handwriting Learning System based on Dynamic
  Movement Primitives (DMPs).
\newblock \emph{arXiv preprint arXiv:2012.03898}.

\bibitem[{Ross, Gordon, and Bagnell(2011)}]{ross2011reduction}
Ross, S.; Gordon, G.; and Bagnell, D. 2011.
\newblock A reduction of imitation learning and structured prediction to
  no-regret online learning.
\newblock In \emph{Proceedings of the fourteenth international conference on
  artificial intelligence and statistics}, 627--635. JMLR Workshop and
  Conference Proceedings.

\bibitem[{Sun, Qian, and Xu(2014)}]{sun2014robot}
Sun, Y.; Qian, H.; and Xu, Y. 2014.
\newblock Robot learns Chinese calligraphy from demonstrations.
\newblock In \emph{2014 IEEE/RSJ International Conference on Intelligent Robots
  and Systems}, 4408--4413. IEEE.

\bibitem[{Suzuki et~al.(2021)Suzuki, Kanamura, Suga, Mori, and
  Ogata}]{suzuki2021air}
Suzuki, K.; Kanamura, M.; Suga, Y.; Mori, H.; and Ogata, T. 2021.
\newblock In-air knotting of rope using dual-arm robot based on deep learning.
\newblock In \emph{2021 IEEE/RSJ International Conference on Intelligent Robots
  and Systems (IROS)}, 6724--6731. IEEE.

\bibitem[{Teo, Burdet, and Lim(2002)}]{teo2002robotic}
Teo, C.~L.; Burdet, E.; and Lim, H. 2002.
\newblock A robotic teacher of Chinese handwriting.
\newblock In \emph{Proceedings 10th Symposium on Haptic Interfaces for Virtual
  Environment and Teleoperator Systems. HAPTICS 2002}, 335--341. IEEE.

\bibitem[{Williams, Toussaint, and Storkey(2006)}]{williams2006extracting}
Williams, B.~H.; Toussaint, M.; and Storkey, A.~J. 2006.
\newblock Extracting motion primitives from natural handwriting data.
\newblock In \emph{International Conference on Artificial Neural Networks},
  634--643. Springer.

\bibitem[{Yin et~al.(2016)Yin, Alves-Oliveira, Melo, Billard, and
  Paiva}]{yin2016synthesizing}
Yin, H.; Alves-Oliveira, P.; Melo, F.~S.; Billard, A.; and Paiva, A. 2016.
\newblock Synthesizing robotic handwriting motion by learning from human
  demonstrations.
\newblock In \emph{Proceedings of the 25th International Joint Conference on
  Artificial Intelligence}, 3530–3537. AAAI Press.
\newblock ISBN 9781577357704.

\bibitem[{Zhang et~al.(2018)Zhang, McCarthy, Jow, Lee, Chen, Goldberg, and
  Abbeel}]{zhang2018deep}
Zhang, T.; McCarthy, Z.; Jow, O.; Lee, D.; Chen, X.; Goldberg, K.; and Abbeel,
  P. 2018.
\newblock Deep imitation learning for complex manipulation tasks from virtual
  reality teleoperation.
\newblock In \emph{2018 IEEE International Conference on Robotics and
  Automation (ICRA)}, 5628--5635. IEEE.

\bibitem[{Zhao et~al.(2019)Zhao, Zheng, Xu, and Wu}]{zhao2019object}
Zhao, Z.-Q.; Zheng, P.; Xu, S.-t.; and Wu, X. 2019.
\newblock Object detection with deep learning: A review.
\newblock \emph{IEEE transactions on neural networks and learning systems},
  30(11): 3212--3232.

\end{thebibliography}
\label{sec:reference_examples}

\end{document}